\documentclass{ieeeaccess}
\usepackage{cite}
\usepackage{amsmath,amssymb,amsfonts}
\usepackage{algorithmic}
\usepackage{graphicx}
\usepackage{textcomp}
\usepackage{hyperref}
\usepackage{booktabs}
\usepackage{subfloat}
\def\BibTeX{{\rm B\kern-.05em{\sc i\kern-.025em b}\kern-.08em
    T\kern-.1667em\lower.7ex\hbox{E}\kern-.125emX}}
\begin{document}
\history{}

\doi{}

\title{LatentColorization: Latent Diffusion-Based Speaker Video Colorization}
\author{
    \uppercase{
    Rory Ward\authorrefmark{1} \authorrefmark{3} \IEEEmembership{Member, IEEE}, 
    Dan Bigioi\authorrefmark{3} \IEEEmembership{Graduate Student Member, IEEE}, 
    Shubhajit Basak\authorrefmark{2} \IEEEmembership{Member, IEEE},
    John G. Breslin\authorrefmark{3} \IEEEmembership{Senior Member, IEEE},
    Peter Corcoran\authorrefmark{3} \IEEEmembership{Fellow, IEEE}
    }
}
\address[1]{SFI Centre for Research Training in Artificial Intelligence, Data Science Institute, University of Galway, University Road, H91 TK33, Ireland.}
\address[2]{School of Computer Science, University of Galway, University Road, Galway, H91 TK33, Ireland}
\address[3]{School of Engineering, University of Galway, University Road, Galway, H91 TK33, Ireland}
\tfootnote{This work was conducted with the financial support of Science Foundation Ireland through the SFI Centre for Research Training in Artificial Intelligence under Grant No. 18/CRT/6223, the Insight SFI Centre for Data Analytics under Grant No. 12/RC/2289\_P2, and the SFI Centre for Research Training in Digitally-Enhanced Reality (d-real) under Grant No. 18/CRT/6224. For the purpose of Open Access, the author has applied a CC BY public copyright licence to any Author Accepted Manuscript version arising from this submission.}

\markboth
{Ward \headeretal: LatentColorization: Latent Diffusion-Based Speaker Video Colorization}
{Ward \headeretal: LatentColorization: Latent Diffusion-Based Speaker Video Colorization}

\corresp{Corresponding author: Rory Ward (e-mail: R.Ward15@nuigalway.ie).}

\begin{abstract}

While current research predominantly focuses on image-based colorization, the domain of video-based colorization remains relatively unexplored. Most existing video colorization techniques operate on a frame-by-frame basis, often overlooking the critical aspect of temporal coherence between successive frames. This approach can result in inconsistencies across frames, leading to undesirable effects like flickering or abrupt color transitions between frames. To address these challenges, we harness the generative capabilities of a fine-tuned latent diffusion model designed specifically for video colorization, introducing a novel solution for achieving temporal consistency in video colorization, as well as demonstrating strong improvements on established image quality metrics compared to other existing methods. Furthermore, we perform a subjective study, where users preferred our approach to the existing state of the art. Our dataset encompasses a combination of conventional datasets and videos from television/movies. In short, by leveraging the power of a fine-tuned latent diffusion-based colorization system with a temporal consistency mechanism, we can improve the performance of automatic video colorization by addressing the challenges of temporal inconsistency. A short demonstration of our results can be seen in some example videos available at \color{blue}\url{https://youtu.be/vDbzsZdFuxM}\color{black}.
\end{abstract}


\titlepgskip=-15pt

\maketitle

\section{Introduction}
\label{sec:intro}

With the rapid increase in the popularity of streaming video in recent years, today's media consumers have become accustomed to high-definition and vibrant video experiences, in color and on demand. However, there are also many substantial video archives with content that remains available in black and white only. Unlocking the potential of these archives, and infusing them with color, presents an exciting opportunity to engage with modern audiences, and breathe new life into classic movies and television episodes. By seamlessly blending cutting-edge technology with classic content, we not only enhance the visual appeal for contemporary viewers but also ensure that the historical significance of these timeless works are faithfully maintained.

\begin{subfigures}
\begin{figure*}
\begin{center}
  \includegraphics[width=0.7\textwidth]{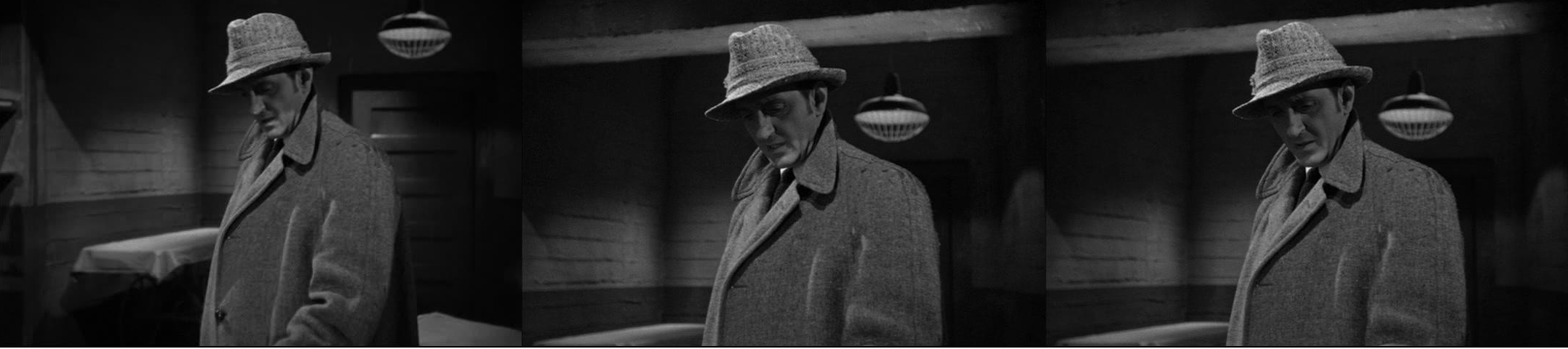}
  \caption{"Sherlock Holmes and the Woman in Green" (1945) black-and-white frames.}
  \label{fig:bandw}
\end{center}
\end{figure*}
\begin{figure*}
\begin{center}
  \includegraphics[width=0.7\textwidth]{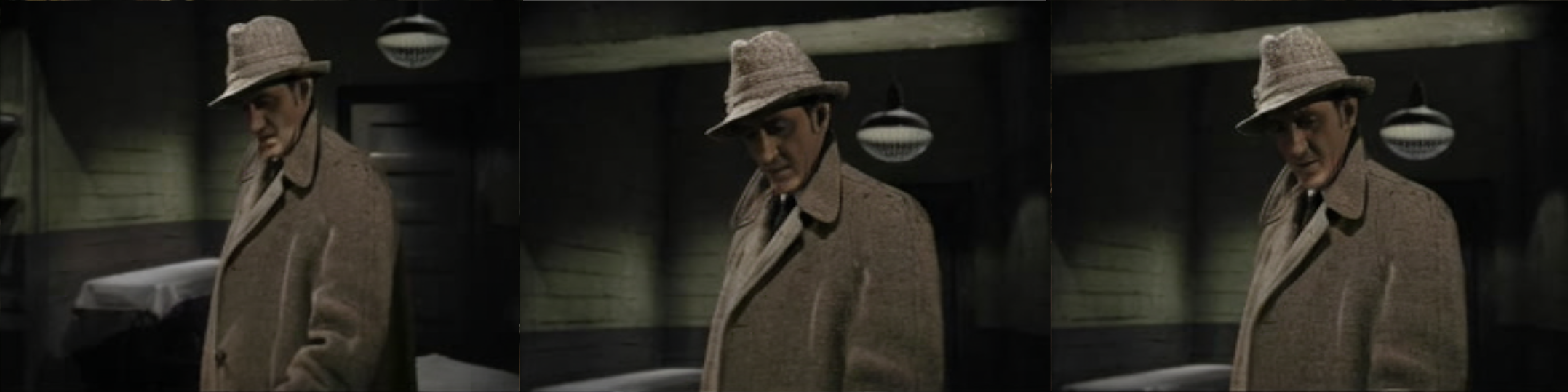}
  \caption{"Sherlock Holmes and the Woman in Green" (1945)  LatentColorization output frames.}
  \label{fig:lc_output}
\end{center}
\end{figure*}
\end{subfigures}

\subsection{Traditional Colorization}

Colorizing black-and-white multimedia is a formidable challenge characterized by its inherent complexity. It presents a `one-to-many' scenario, wherein multiple feasible colorization outcomes can be derived for a single black-and-white video, as illustrated by recent research \cite{https://doi.org/10.48550/arxiv.2005.10825}.

Traditional approaches for video colorization are manual and labor-intensive, demanding the dedicated efforts of interdisciplinary teams comprised of skilled colorists and rotoscoping animators, artists and historians. These teams invest extensive hours to ensure the production of a convincing and coherent end result. The intricacies of colorization are particularly difficult in the realm of videos, where the sheer volume of frames per second amplifies the complexity \cite{Pierre2021}. Therefore, automation of the video colorization process is highly desirable. 

\subsection{Automatic Colorization}

Automatic video colorization can be seen as a means to significantly reduce the cost traditionally associated with manually colorizing/restoring vintage movies, an expensive proposition that is often limited to organizations with substantial budgets. Since the labor costs associated with expert colorists are a significant barrier, manual colorization has also been largely limited to popular films or TV shows (e.g., Doctor Who), with numerous other works (social history movies, documentaries, films by lesser-known directors, etc.) omitted where the cost-benefit analysis could not justify their colorization.

As a consequence, various research efforts have tackled the need to automate aspects of the colorization process. These efforts span from earlier methods such as histogram matching \cite{LIU20121673}, to more recent interactive approaches such as scribble-based systems \cite{scribble} and exemplar-based approaches \cite{8954242}, as well as more recent developments in terms of deep learning-based colorization \cite{antic2019deoldify}. While the results still lag behind those that can be achieved of an experienced human colorizer, the automated approaches referred to above have made significant advancements in terms of their accuracy over prior systems.

In terms of the state-of-the-art, one current benchmark for automatic video colorization is held by Wan et al. \cite{https://doi.org/10.48550/arxiv.2203.17276}. However, it is important to note that their approach not only colorizes but also restores videos, making it a difficult benchmark for systems that are focused solely on colorization. DeOldify \cite{antic2019deoldify}, provides colorized outputs without image restoration, and therefore can be more easily compared against colorization-only approaches such as the one presented in this paper.

Recent research \cite{https://doi.org/10.48550/arxiv.1806.09594} has shown the advantages of self-supervised learning methodologies for colorization, removing the resource-intensive need for creating and curating manually labelled datasets for training models. Constructing custom labelled datasets can be a resource-intensive and time-consuming endeavor, particularly when dealing with video content which has both static- and motion-related information.

\subsection{Research Contribution}

Driven by the recent increase in the adoption of diffusion models \cite{sohl2015deep, ho2020denoising, https://doi.org/10.48550/arxiv.2105.05233}, the field of generative modelling has produced a variety of contributions including Stable Diffusion \cite{nokey}, Imagen \cite{https://doi.org/10.48550/arxiv.2205.11487}, and DALL•E 2 \cite{https://doi.org/10.48550/arxiv.2204.06125} which have gained attention in both research and the mainstream media.

Within the context of video colorization, the majority of techniques are based on GAN-based methods \cite{https://doi.org/10.48550/arxiv.1406.2661, https://doi.org/10.48550/arxiv.1906.09909}, as well as the utilization of transformer-based approaches \cite{https://doi.org/10.48550/arxiv.1706.03762} such as those featured in \cite{https://doi.org/10.48550/arxiv.2102.04432, DBLP:journals/corr/abs-2109-02614, https://doi.org/10.48550/arxiv.2203.17276}. Notably, Saharia et al. \cite{saharia2022palette} propose leveraging diffusion models for various image-to-image tasks, including colorization. 

This paper introduces an innovative approach to video-based colorization, employing a latent-based denoising diffusion model. Our method demonstrates improvements over the state-of-the-art DeOldify \cite{antic2019deoldify} method, across a range of standard evaluation metrics including Power Signal to Noise Ratio (PSNR), Structural Similarity (SSIM), Fréchet Inception Distance (FID), Fréchet Video Distance (FVD), and Naturalness Image Quality Evaluator (NIQE). Furthermore, we provide comparative results for Blind/Referenceless Image Spatial Quality Evaluator (BRISQUE). It is also worth noting that our method yields an average improvement of approximately 18\% when FVD is employed as the evaluation metric. This result is also collaborated by our user study where LatentColorization is preferred 80\% of the time to the previous state-of-the-art.

We introduce a novel system for achieving temporal consistency in video colorization through the application of a latent diffusion model. A sample visual, before and after, is given in Figures \ref{fig:bandw} and \ref{fig:lc_output}.

\begin{figure*}[h]
  \centering
  \includegraphics[width=0.9\linewidth]{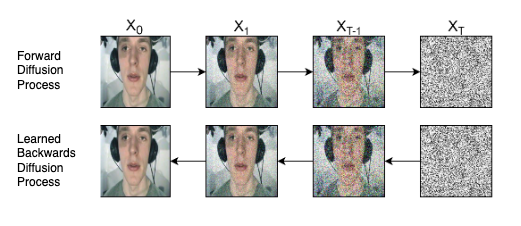}
  \caption{Diagram of the Diffusion Process: This diagram illustrates the operation of the diffusion model in both the forward and backward processes. In the forward process, it visually portrays the incremental addition of Gaussian noise to the input image $x_0$ until it becomes visually indistinguishable from Gaussian noise $x_T$ (top). Subsequently, it showcases the learned backward diffusion process, where the model gradually removes the Gaussian noise from $x_T$ to return to the original image $x_0$ (bottom).}
  \label{fig:DiffusionModel}
\end{figure*}

To summarise, the unique contributions of our proposed work are as follows:

\begin{itemize}
  \item We apply fine-tuned latent diffusion models to the automatic video colorization task with exemplar frame conditioning.
  
  \item We ensure temporal consistency in automatic video colorization through the use of our autoregressive conditioning mechanism.
  
  \item We build a novel end-to-end video colorization application. 

    \item We achieve state-of-the-art performance over a range of datasets, metrics and evaluations.
    
\end{itemize}

The structure of this paper is as follows: In \hyperref[sec:related_work]{§2}, we examine related work. \hyperref[sec:methodology]{§3} provides an in-depth description of our methodology. Then, \hyperref[sec:evaluation]{§4} presents the results of our evaluations, which are further examined in \hyperref[sec:discussion]{§5}. Conclusions are given in \hyperref[sec:conclusion]{§6}, and we outline our future research directions in \hyperref[sec:future_work]{§7}.

\section{Related Work}
\label{sec:related_work}

\subsection{Conventional Deep Learning Approaches}

Generative adversarial networks, commonly referred to as GANs \cite{https://doi.org/10.48550/arxiv.1406.2661}, have emerged as a common technology in the enhancement of existing video content, in domains including sign-language addition \cite{9905589}, low-light enhancement \cite{9609683}, and video colorization \cite{8954242}. GAN-based methods have also been extensively used for image colorization \cite{https://doi.org/10.48550/arxiv.1611.07004, 10.1145/3355089.3356561, 10.1145/3272127.3275090, https://doi.org/10.48550/arxiv.1810.05399, https://doi.org/10.48550/arxiv.1801.02753, https://doi.org/10.48550/arxiv.1706.06918, 2021, https://doi.org/10.48550/arxiv.1702.06674}. For example,  Isola et al. proposed Pix2Pix \cite{https://doi.org/10.48550/arxiv.1611.07004}, which has performed well on various benchmarks, including the FID-5K benchmark using the ImageNet Val dataset. In the context of video colorization, DeOldify \cite{antic2019deoldify} and more recently GCP \cite{wu2022vivid} stand out as two of the more prominent GAN-based approaches. 

DeOldify \cite{antic2019deoldify} is a self-attention-based Generative Adversarial Network (GAN) \cite{https://doi.org/10.48550/arxiv.1805.08318}. It incorporates NoGAN training \cite{nogan} and adheres to a Two Time Scale Update Rule \cite{https://doi.org/10.48550/arxiv.1706.08500}. While DeOldify is capable of generating credible colorizations, it has a tendency to produce somewhat subdued or less vibrant colors, characteristic of GAN-based systems. 

GCP \cite{wu2022vivid} leverages color priors encapsulated in a pretrained Generative Adversarial Networks (GAN) for automatic colorization. Specifically, they “retrieve” matched features (similar to exemplars) via a GAN encoder and then incorporate these features into the colorization process with feature modulations.

Other works, such as \cite{https://doi.org/10.48550/arxiv.1905.03023, 9233962, https://doi.org/10.48550/arxiv.2011.12528}, have also made contributions to the field of video colorization. It is important to note that GANs, due to their reliance on multiple loss functions, are challenging to train, susceptible to mode collapse, and often encounter convergence issues \cite{NIPS2017_44a2e080, https://doi.org/10.48550/arxiv.1612.02136, https://doi.org/10.48550/arxiv.1606.03498}. Furthermore, only certain GAN-based automatic colorization systems consider temporal consistency, such as Zhao et al. \cite{Zhao_2023}. This means that the systems that do not account for temporal consistency do not maintain coherence across successive frames, which is a crucial aspect of video colorization.

Video Colorization with Hybrid Generative Adversarial Network (VCGAN) \cite{Zhao_2023} is an end-to-end recurrent colourization network that prioritises temporal consistency in automatic video colorization.

DeepRemaster, as introduced by Iizuka et al. in their work \cite{IizukaSIGGRAPHASIA2019}, is a Convolutional Neural Network (CNN)-based colorization system. As well as colorization, it also performs super-resolution, noise reduction, and contrast enhancement. Its performance makes it a suitable benchmark for comparison in our work.

Transformers, known for their success in diverse machine learning domains, including Natural Language Processing (NLP) and Computer Vision (CV), have achieved state-of-the-art results in various low-resolution computer vision tasks, exemplified by their second-place ranking on the FID-5K benchmark using the ImageNet Val dataset. However, the computational complexity of their self-attention mechanism scales significantly with higher image resolutions, presenting a challenge for handling high-resolution images \cite{https://doi.org/10.48550/arxiv.2103.14031, DBLP:journals/corr/abs-2109-02614}. While ongoing research efforts aim to mitigate this challenge, it remains an open area of investigation. Unlike GANs, transformers exhibit greater resilience to mode collapse, thanks to their distinctive attention mechanism.

Kumar et al. have introduced ColTran \cite{https://doi.org/10.48550/arxiv.2102.04432}, a transformer-based image colorization model that operates through a three-step process. Initially, it colorizes a low-resolution version of the image, as it leverages self-attention, which is computationally demanding for high-resolution photos. Subsequently, it upscales the image and then the colors, yielding high-resolution colorized images. ColTran excels in producing vibrant colorizations, yet it falls short of catering to the specific demands of video colorization, leading to inconsistencies in video colorizations.

\subsection{Diffusion Models}

Diffusion models, as initially introduced by Sohl-Dickstein et al. \cite{sohl2015deep}, operate by learning how to reconstruct data from noise. They encompass two distinctive stages:

Forward Diffusion Process: In this phase, Gaussian noise is incrementally incorporated into the data through a stepwise progression spanning multiple timesteps. This gradual introduction of noise gradually transforms the original information until the desired level of diffusion or alteration is attained.

Reverse Diffusion Process: Subsequently, a learning model is employed to reverse this diffusion process, effectively reconstructing the original data \cite{https://doi.org/10.48550/arxiv.2006.11239}, as illustrated in Figure~\ref{fig:DiffusionModel}. Unlike Generative Adversarial Networks (GANs), diffusion models are resilient to mode collapse, and they have demonstrated success across various domains, including video generation \cite{https://doi.org/10.48550/arxiv.2210.02303, anonymous2023phenaki}, audio generation \cite{https://doi.org/10.48550/arxiv.2009.09761, https://doi.org/10.48550/arxiv.2207.09983}, and image generation \cite{https://doi.org/10.48550/arxiv.2210.02303, nokey, https://doi.org/10.48550/arxiv.2204.06125}.

An illustration of the application of diffusion models to still-image colorization can be found in Palette \cite{https://doi.org/10.48550/arxiv.2111.05826}, a diffusion model tailored for a variety of image-to-image tasks. Palette attains the top position on the leader-board in the FID-5K benchmark using the ImageNet Val dataset.

Concurrently, Liu et al. \cite{liu2023video} are engaged in research focused on the challenge of achieving temporally consistent video colorization, employing pre-trained diffusion models. A distinction lies in their approach as they utilize text-based conditioning for their system. In contrast, our methodology relies on exemplar frames as the conditioning input. This strategic choice was made based on our belief that using an image for conditioning provides a higher degree of expressive control compared to text-based approaches.

A challenge with diffusion models is their demanding computational requirements during both the training and testing phases. Nevertheless, ongoing research endeavors are actively addressing this issue \cite{https://doi.org/10.48550/arxiv.2106.05931, https://doi.org/10.48550/arxiv.2112.07068, https://doi.org/10.48550/arxiv.2112.07804}. Several approaches have emerged to mitigate this challenge:

\textbf{Down-sampling and Super-resolution:} Works such as Make-A-Video \cite{singer2022makeavideo} tackle this issue by initially down-sampling the resolution of images in the diffusion process and subsequently restoring the resolution using a super-resolution algorithm.

\textbf{Latent Diffusion:} Another approach, exemplified by Latent Diffusion \cite{nokey}, modifies the diffusion process to operate in the latent space of a trained autoencoder, as opposed to the pixel space. This results in reductions in both inference and training times due to the reduced dimensionality of the data inputted into the diffusion process.


The work presented in this paper represents the first instance, where the video colorization task is tackled through the use of an image-to-image latent diffusion model employing exemplar frames.

\section{Methodology}
\label{sec:methodology}

\subsection{Design Considerations}

For achieving temporally consistent video colorization, there are two popular methods:

\textbf{Implicit Temporal Consistency:} In this approach, the notion of ensuring explicit temporal consistency is considered unnecessary. The belief is that with a sufficiently accurate system and reasonably similar input (e.g., consecutive frames in a video sequence), the colorized output should naturally exhibit similarity and relative consistency. As a result, temporal consistency is managed implicitly.

\textbf{Explicit Temporal Consistency:} This project aligns with the second methodology, which emphasizes the explicit addressing of temporal consistency. Rather than relying on the system to learn it implicitly, this approach involves conditioning for temporal consistency explicitly. The advantages of this approach include reduced training time, decreased data requirements, and a lower computational load. However, it necessitates more intricate system engineering to explicitly convey the requirements to the system.

Within the realm of implicit temporal consistency methodologies, several approaches are prevalent, with three of the most common being:

\textbf{Optical Flow-Based:} Optical flow-based colorization methods operate by conditioning the system to maintain color consistency over time. However, it is worth noting that a limitation of this approach is the potentially high computational cost associated with calculating optical flow, making it less practical in certain applications \cite{10.1117/12.2037496}.

\textbf{Exemplar-Based:} Exemplar-based methods involve providing the system with a reference image to guide its colorization process. This typically entails human intervention or a database retrieval algorithm with a collection of reference images \cite{10.1145/1409060.1409105}.

\textbf{Hybrid-Based:} Some methods adopt a hybrid approach by combining different methodologies to harness the benefits of multiple systems simultaneously. This strategy, as seen in works like \cite{imvi2022,8954242}, seeks to leverage the strengths of various techniques to enhance overall performance.

\begin{figure}[h]
  \centering
  \includegraphics[width=0.8\linewidth]{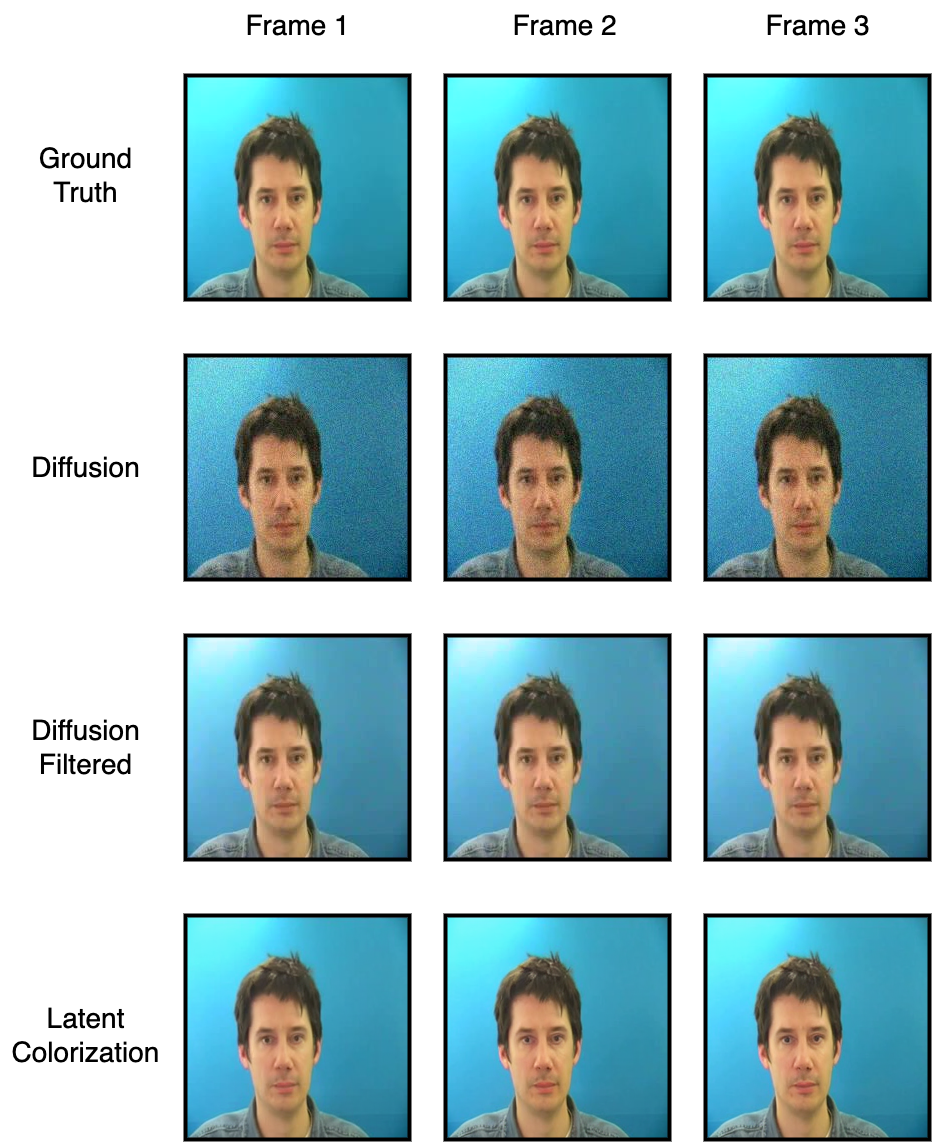}
  \caption{Comparison of 3 consecutive frames with different operations applied: First Row (Ground Truth): This row showcases the original, unaltered images, representing the ground truth reference. Second Row (Diffusion Model): In the second row, you can observe the colorization output generated by our original diffusion model. Third Row (Diffusion Model with Post-Processing): Here, the output of the diffusion model is presented with an additional post-processing procedure applied to enhance the results. Fourth Row (LatentColorization): The final row displays the results obtained from  LatentColorization .}
  \label{fig:Diffusion_Comparisons}
\end{figure}

\subsection{Data Processing}

We use the following datasets as part of our experiments:

GRID Dataset: The GRID dataset \cite{cooke_martin_2006_3625687} is a collection of video recordings featuring individuals speaking. It encompasses high-quality facial recordings of 1,000 sentences spoken by each of 34 talkers, with a distribution of 18 males and 16 females, resulting in a total of 34,000 sentences.

Lombard Grid Dataset: An extension of the GRID dataset, the Lombard Grid dataset \cite{lombardGrid}, includes 54 talkers, each contributing 100 utterances. Among these 54 talkers, 30 are female, and 24 are male, expanding the dataset's diversity.

Sherlock Holmes Movies Dataset: This dataset is a collection of professionally colorized frames extracted from 'Sherlock Holmes and the Woman in Green,' 'Sherlock Holmes Dressed to Kill,' 'Sherlock Terror by Night,' and 'Sherlock Holmes and the Secret Weapon.'

These diverse datasets provide a foundation for our research in the field of speaker video colorization and temporally consistent diffusion models.

Our dataset consisted of 10,000 frames allocated for training the model, with an additional 700 frames reserved for testing purposes. Each frame was uniformly resized to 128x128 pixels.

To ensure the generalizability of our model, the training and testing frames were derived from distinct subjects, mitigating the risk of artificially inflated performance measures that would not extend to real-world scenarios.

By conducting tests on benchmark datasets, we were able to compare our approach against previous methods. Furthermore, testing on the Sherlock Holmes-related data provided a valuable means of comparing our results to expert human colorizations. Additionally, training on open-domain videos underscores the potential of these resources in advancing the field of video colorization.

\begin{subfigures}
\begin{figure*}[h]
  \centering
  \includegraphics[width=1\linewidth]{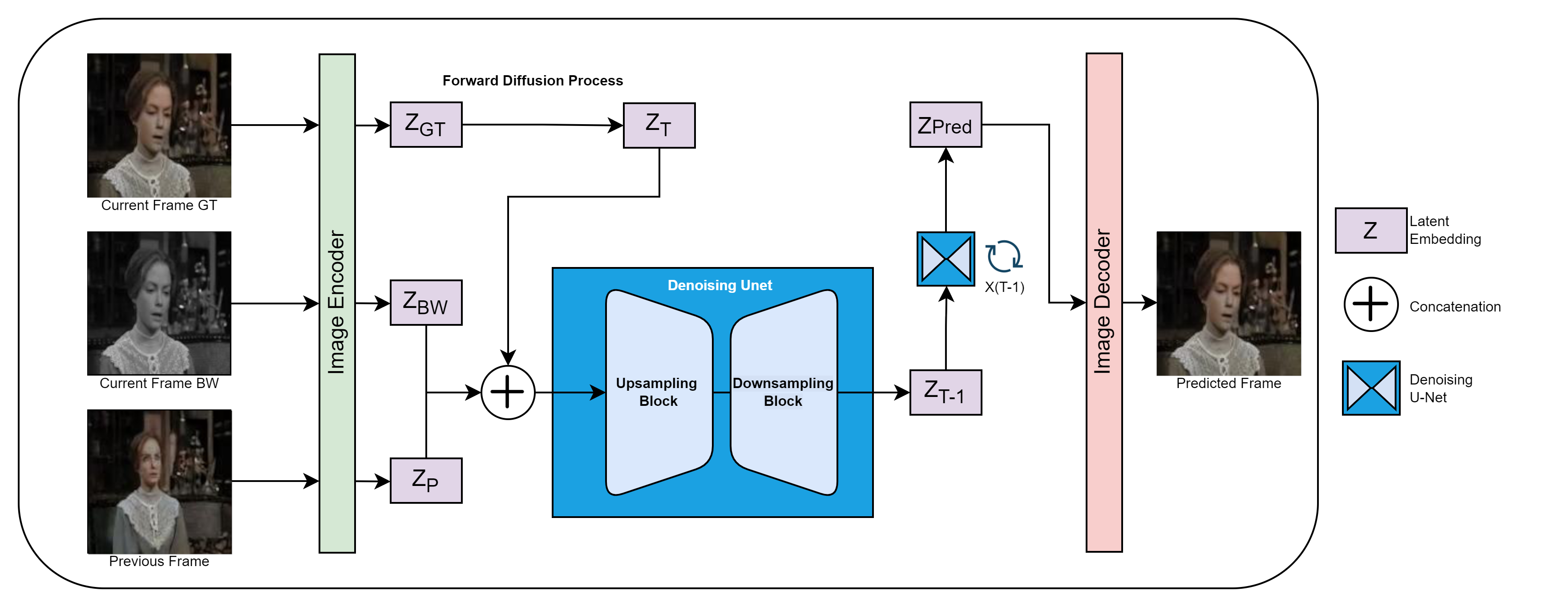}
  \caption{The system architecture during training is depicted in the diagram, illustrating the key elements of the network and their interactions: Image Encoder: This component is responsible for encoding the input frames into embedding representations. It generates the ground truth embedding $Z_{GT}$, the embedding of the current black-and-white frame $Z_{BW}$, and the embedding of the previous color frame $Z_P$. Denoising Unet: This is a critical part of the architecture, responsible for denoising and refining the embeddings generated by the Image Encoder that have passed through the forward diffusion process. Conditioning Mechanism: The conditioning mechanism is integral to the network, providing contextual information and conditioning signals to guide the colorization process. It takes into account various embeddings, including $Z_{BW}$, $Z_P$, and $Z_{T}$, which represent the black and white input frame, the output of the model at the previous timestep, and the noisy frame to be denoised. Image Decoder: This component is responsible for decoding the predicted frames from their embedding representations. The architecture's design and interactions are essential for the model's training process, ensuring that it learns to generate accurate and temporally consistent colorizations over multiple timesteps.}
  \label{fig:SystemArchitecture}
\end{figure*}
\begin{figure*}[h]
  \centering
  \includegraphics[width=1\linewidth]{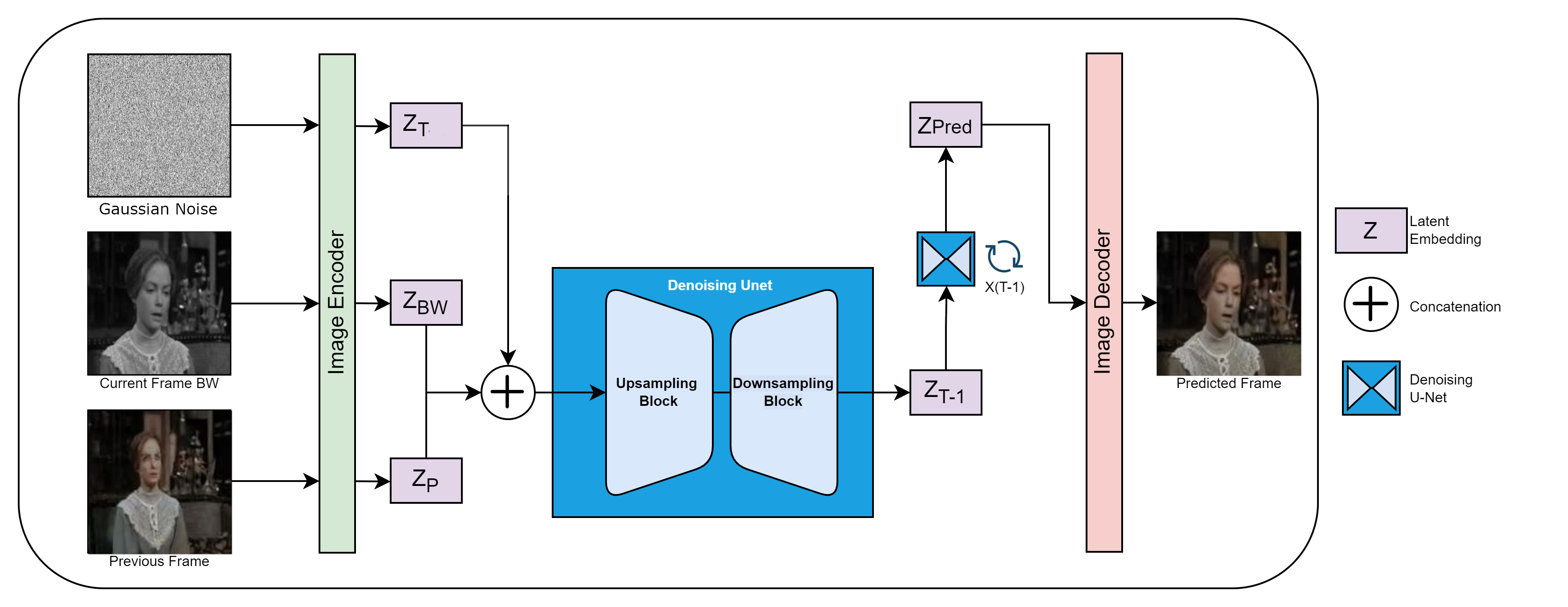}
  \caption{During inference, the system architecture remains largely consistent with the training phase, with one significant difference: Gaussian Noise in Place of Ground Truth Frame: Instead of the ground truth frame, the system introduces Gaussian noise as input during the testing phase. This alteration simulates real-world scenarios where the model must colorize frames without the ground truth. The rest of the architecture, including the Image Encoder, Denoising Unet, Conditioning Mechanism, Image Decoder, and their interactions, remains unchanged. This design allows the model to assess its performance under conditions that more closely resemble practical, ground truth-free scenarios.}
  \label{fig:SystemArchitecture_Testing}
\end{figure*}
\end{subfigures}

\subsection{System Overview}

\subsubsection{Image Diffusion Based Set Up}
In our initial exploration, we considered adopting a setup akin to Palette \cite{https://doi.org/10.48550/arxiv.2111.05826}, incorporating our temporal consistency mechanism and initial frame biasing, which will be elaborated on in Section \ref{subsec:temp_cons}. However, we observed sub-optimal performance from this configuration, as the system's outputs exhibited undesired residual speckled noise, as illustrated in Fig.~\ref{fig:Diffusion_Comparisons}.

To address the speckled noise in the diffusion colorization outputs, we explored two approaches:

\textbf{Non-Linear Means (nlmeans) Clustering:} We initially applied the nlmeans clustering algorithm \cite{nlmeans} to the images to mitigate the noise. However, this method relies on a hyper-parameter that dictates the filter's strength. A stronger filter results in smoother images but may inadvertently remove high-quality details, such as hair and facial features. Conversely, a weaker filter may leave more residual speckled noise unfiltered.

\textbf{Overlaying Colorized Output with Black-and-White Inputs:} As an alternative, we experimented with overlaying the colorized output with the original black-and-white inputs. This approach yielded superior results compared to the nlmeans filter, and it required less parameter tuning for filter strength. We opted to proceed with this approach, referred to as 'Diffusion Filtered'.

Despite our efforts to optimize noise reduction while preserving critical details, the final output quality still fell short of our improved approach, LatentColorization, which we will detail in the following section. Consequently, our final experiments did not incorporate the Palette-based approach \cite{https://doi.org/10.48550/arxiv.2111.05826}.

\subsubsection{Latent Diffusion Based Set Up}

Inspired by Latent Diffusion \cite{nokey}, we devised LatentColorization.

LatentColorization comprises three core components: an autoencoder, a latent diffusion model, and a conditioning mechanism, as visually represented in Fig~\ref{fig:SystemArchitecture}.

The latent diffusion model follows a two-step process, commencing with the forward diffusion phase (formulated in Eqn.~\ref{Forward_Diffusion}). During this phase, Gaussian noise is systematically introduced to the data, incrementally transforming it until it becomes indistinguishable from Gaussian noise. During the second phase,  the learned backward diffusion process is applied. This is where a neural network is trained to learn the original data distribution, and to draw samples from it by reconstructing the data from Gaussian noise. We represent formulations of this process with conditioning in Eqn.\ref{conditioned_backwards_diffusion} and without conditioning, as in Eqn.\ref{backwards_diffusion}.

The forward diffusion process, as defined by \cite{sohl2015deep}, can be represented by the following formula:

\begin{equation}
    q(x_t|x_{t-1}) = N(x_t; \mu_t=\sqrt{1-\beta}x_{t-1},\Sigma_t = \beta_tI)
    \label{Forward_Diffusion}
\end{equation}

In this formulation, the probability distribution $q(\cdot)$ of the image at each timestep $x_t$, given the previous timestep $x_{t-1}$, is characterized as a normal distribution $N$. This distribution is centred around a mean equal to the previous timestep $x_{t-1}$, with noise incorporated. The magnitude of this noise is determined by the noise scheduler $\beta$ at time $t$ and is further modulated by the identity matrix $I$. The noise scheduler $\beta$ typically follows a linear pattern, as exemplified in \cite{https://doi.org/10.48550/arxiv.2006.11239}, or a cosine pattern, as demonstrated in \cite{nichol2021improved}.

The backward diffusion process, in accordance with \cite{ho2020denoising}, can be defined as follows:

\begin{equation}
    p_\theta(x_{t-1}|x_{t}) = N(x_{t-1}; \mu_\theta(x_t,t),\Sigma_\theta(x_t,t))
    \label{backwards_diffusion}
\end{equation}

In this definition, the probability distribution $p_\theta(\cdot)$ of the slightly denoised image $x_{t-1}$, given the noisier image $x_t$, is characterized as a normal distribution $N()$. This distribution has a mean denoted as $\mu$ and a variance represented by $\Sigma$, both of which are learned and parameterized by the neural network indicated by $\theta$.

The diffusion process can be conditioned using the following equation:

\begin{equation}
    p_\theta(x_{0:T}|y) = p_\theta(x_t)\prod_{t=1}^{T}p_\theta (x_{t-1}|x_t,y)
    \label{conditioned_backwards_diffusion}
\end{equation}

In this equation, the probability density function $p_\theta$ is akin to the unconditioned diffusion process, but conditioning is introduced at each timestep of the diffusion process, denoted as $p_\theta(x_{t-1}|x_t,y)$. In our specific scenario, the conditions encompass the previous frame, the grayscale frame, and the current frame during training, as illustrated in Fig.\ref{fig:SystemArchitecture}. During inference, the conditions consist of the previous frame, the grayscale frame, and noise, as indicated in Fig.\ref{fig:SystemArchitecture_Testing}.

For a visual representation of our network architecture during training and inference, as well as a breakdown of where each equation is utilized, please refer to Fig~\ref{fig:SystemArchitecture} and Fig~\ref{fig:SystemArchitecture_Testing}. Additionally, for a more in-depth explanation of these equations and their derivation, you can explore the references provided in \cite{ho2020denoising,sohl2015deep,song2019generative}.

In the training process, the current frame ground truth, the current frame in black and white, and the previous frame are fed into the image encoder. These images are compressed into their respective embeddings, namely $Z_{GT}$, $Z_{BW}$, and $Z_P$. The chosen autoencoder for this purpose is a Vector Quantized Variational AutoEncoder (VQ-VAE), as detailed in \cite{https://doi.org/10.48550/arxiv.1711.00937}.

During the forward diffusion process, the current frame's ground truth embedding $Z_{GT}$ has noise applied to it based on the noise timestep, resulting in $Z_T$. Simultaneously, the ground truth black and white embedding $Z_{BW}$ and the previous frame embedding $Z_P$ are concatenated. The noised embedding $Z_T$ is then denoised using the Unet and conditioned on $Z_{BW}$ and $Z_P$.

During the backward diffusion process, the neural network learns to predict the noise that was added during the forward diffusion process at time step $T$. Denoising the noise embedding $Z_{T}$ using the predicted noise results in $Z_{T-1}$. We use a simple mean square error loss between the predicted noise, vs the actual noise added to the embedding in order to train the network. 

By employing the previous frame as conditioning, temporal consistency between frames is ensured throughout the video sequence, resulting in coherent colorization.

During inference, the same process as the training scheme is followed, with the exception that the model is fed pure Gaussian noise representing frame $Z_T$. The denoising is then repeated $T$ times, after which the denoised embedding is passed through to the image decoder in order to produce the predicted frame. This process is depicted explicitly in Fig.\ref{fig:SystemArchitecture_Testing}. 

The system can be used in two different ways. First, it can be employed in an entirely end-to-end manner, where no additional guidance from the user is needed. In this setup, the system serves as an image colorization tool for the first frame. Then, this initial colorized frame is used in an auto-regressive fashion to guide the colorization of subsequent frames in the video clip. Second, the system can be used interactively, allowing the user to manually colorize the initial frame. This manual colorization becomes the condition for initiating the colorization of the following frames. This second approach provides control over the colorization process but requires the user to provide the initial colorization.

\subsection{Temporal Consistency}
\label{subsec:temp_cons}

Temporal consistency was maintained through an auto-regressive conditioning mechanism, where the current video frame was conditioned on the previous frame and the grayscale version of the current frame. This approach ensured that colorization remained consistent across the video frames. For a detailed illustration, refer to Fig~\ref{fig:SystemArchitecture} and Eqn~\ref{eqn:Temporal_Consistency}. This mechanism is similar to the approaches used in other studies such as \cite{https://doi.org/10.48550/arxiv.2301.04474} and \cite{https://doi.org/10.48550/arxiv.2301.03396}, where models were conditioned with information from the previous frame to guarantee temporal consistency in the context of video generation. Essentially, maintaining consistent colors throughout a video sequence becomes more achievable when the model can "remember" the colors from the previous frame.

\begin{equation}
\label{eqn:Temporal_Consistency}
C_t = f(C_{t-1..n}, G_t) \forall  t\in T
\end{equation}

In Eqn~\ref{eqn:Temporal_Consistency}, we denote the color image as $C \subseteq \mathbb{R}^{L, a, b}$, the grayscale image as $G \subseteq \mathbb{R}^{L}$, and $f()$ represents the colorization function performed by the neural network. Here, $n$ signifies the length of the conditioning window frame, $T$ is the total length of the video, and $t$ indicates a specific moment within the video sequence. This equation describes how the colorization process is conditioned on both the previous frame and the grayscale version of the current frame, ensuring temporal consistency across the video frames.

Throughout the video sequence, we maintain temporal consistency by providing the colorizer with the previous frame as a reference. However, a challenge arises at the beginning of the sequence, denoted as $t_0$, where there is no previous frame available for conditioning. To address this, we introduce an initial colorized frame at $t_0$. This initial frame is advantageous because it introduces an element of user preference, which can be highly practical. It effectively reduces the video colorization task to that of coloring a single image, which then serves as the starting point for colorizing the entire video with a bias towards the initial frame.

This approach offers flexibility and aligns with human-centric AI concepts for video colorization. We refer to this approach as "initial frame biasing". Additionally, it provides a clear method for evaluating the system, as ground truth is available for the initial frame, making traditional reference-based metrics such as PSNR, SSIM, FID, and FVD effective for assessment. It also allows for a user study where one can compare performance against the ground truth.

\subsection{Hyperparemeter and Training Set Up}
The hyperparameters used in the experiment are detailed in Table \ref{table:hyperparameters}. The experiment employed the ADAM optimizer \cite{https://doi.org/10.48550/arxiv.1412.6980}, with most of the values being adopted from the specifications of Stable Diffusion \cite{nokey}. Any additional hyperparameters were determined through a process of empirical testing.

An image size of 128x128 pixels required a 4x decrease in processing time as opposed to 256x256 pixels. Training at 256x256 pixels takes 165 minutes per epoch on an NVIDIA RTX 2080, whereas training at 128x128 pixels takes 38 minutes per epoch. Using 200 Diffusion steps for training and 50 for testing resulted in good performance. Input channels must be nine to account for the conditioning, three channels for color previous frame, three channels for the image and three channels for the black-and-white current frame. Having a batch size of 256 and a learning rate of $1.25e^{-7}$ resulted in convergence and reasonably fast training times.

\begin{table}[h]
  \centering
  \begin{tabular}{ccc}
  \toprule
     & Train & Test\\
    \midrule
    Image Size & $128x128$ & $128x128$\\
    Total Frames & $10000$ & $700$\\
    Diffusion Steps & $200$ & $50$\\
    Noise Schedule & $Linear$ & $Linear$\\
    Linear Start & $1.5e-03$ & $1.5e-03$\\
    Linear End & $0.0195$ & $0.0195$\\
    Input Channels & $9$ & $9$\\
    Inner Channels & $64$ & $64$\\
    Channels Multiple & $1,2,3,4$ & $1,2,3,4$\\
    Res Blocks & $2$ & $2$\\
    Head Channels & $32$ & $32$\\
    Drop Out & $0$ & $0$\\ 
    Batch Size & $256$ & $8$\\
    Epochs & $350$ & -\\
    Learning Rate & $1.25e^{-7}$ & -\\
    \bottomrule
  \end{tabular}
  \caption{The hyperparameter setup provides the values used for both training and testing}
  \label{table:hyperparameters}
\end{table}	

\section{Evaluation}
\label{sec:evaluation}
The performance evaluation of the colorization process combines both qualitative and quantitative assessments to gauge its success. Following similar colorization studies \cite{https://doi.org/10.48550/arxiv.2203.17276} our work is compared on standard metrics. The key metrics used for this evaluation are as follows:

\vspace{5mm}

\textbf{Power Signal to Noise Ratio (PSNR):} This metric measures the quality of colorized images by comparing them to the corresponding ground truth images. It quantifies the difference between the pixel values of the colorized and ground truth images. Higher PSNR values indicate better performance.

\vspace{5mm}

\textbf{Structural Similarity Index (SSIM):} SSIM evaluates the structural similarity between colorized images and ground truth images. It considers not only pixel values but also the structure and patterns in the images. Higher SSIM values indicate greater similarity to the ground truth.

\vspace{5mm}

\textbf{Fréchet Inception Distance (FID):} FID assesses the distance between the distribution of features extracted from colorized images and real images. Lower FID values indicate closer similarity to real images.

\vspace{5mm}

\textbf{Fréchet Video Distance (FVD):} FVD is a video-specific metric that measures the difference between generated and real videos by comparing the mean and covariance of their features. Lower FVD values represent better video colorization quality.

\vspace{5mm}

\textbf{Naturalness Image Quality Evaluator (NIQE):} NIQE is a referenceless metric that quantifies the naturalness of colorized images using statistical measures. Lower NIQE values indicate more natural-looking images.

\vspace{5mm}

\textbf{Blind/Referenceless Image Spatial Quality Evaluator (BRISQUE):} BRISQUE is another referenceless metric that evaluates the quality of colorized images. It learns the characteristics of natural images and quantifies the deviation from these characteristics. Lower BRISQUE values represent better image quality.

\vspace{5mm}

\textbf{Mean Opinion Score (MOS):} MOS is a weighted average of survey participants' perceived quality of an image or video. Higher MOS score represents a higher opinion of the subjective quality of the media.

\vspace{5mm}

A combination of these quantitative metrics and visual inspection, see Fig~\ref{fig:Comparisons}, allows for a comprehensive assessment of the colorization process, enabling objective and subjective evaluation of its performance. 

Evaluating colorization is a very subjective task, and therefore, as well as the metrics used, a survey was conducted to obtain a subjective measure of our performance. This survey was conducted in a similar manner to the survey conducted by Wu et al. \cite{wu2022vivid}.

\vspace{5mm}

\begin{table*}[t]
\centering
\begin{tabular}{cccccccc}
\\ 
\cmidrule(r){1-8}
                    
Dataset & Method              & PSNR ↑ & SSIM ↑ & FID ↓ & FVD ↓ & NIQE ↓ &  BRISQUE ↓ \\ \hline
Grid \cite{cooke_martin_2006_3625687}& DeOldify \cite{antic2019deoldify}            & 28.07  & 0.79   & 52.67  & 520.75 & 44.04 & \textbf{32.47}  \\
& DeepRemaster \cite{IizukaSIGGRAPHASIA2019}        & 27.7  & 0.77   & 108.68 & 927.91 & 51.19 & 41.44\\
  & ColTran \cite{https://doi.org/10.48550/arxiv.2102.04432}            & 28.08  & 0.76    & 91.76 & 759.32 & 49.69 & 34.55  \\
& GCP \cite{wu2022vivid} & 27.74 & 0.75 & 109.75 & 1555.53 & 48.44 & 33.76 \\
& VCGAN \cite{Zhao_2023} & 27.86 & 0.83 & 67.79 & 951.28 & 44.24 &  37.16 \\
& 
LatentColorization w/o TC & 29.63 & 0.89  & \textbf{20.92}   & 350.35 & 46.4 &  33.73             \\ 
& 
LatentColorization & \textbf{30.88} & \textbf{0.9}  & 22.26   & \textbf{241.94} & \textbf{41.46} & 34.68              \\ \hline  
Lombard Grid \cite{lombardGrid} & DeOldify \cite{antic2019deoldify}            &  30.69  & 0.93   & 17.63  & 396.2 & 46.43 & 33.1  \\
& DeepRemaster \cite{IizukaSIGGRAPHASIA2019}        & 30.09  & 0.95   & 32.9 & 1382.56 & 52.36 & 35.97\\
& ColTran \cite{https://doi.org/10.48550/arxiv.2102.04432}            & 29.96  & 0.89    & 37.7 & 1583.94 & 51.25 & \textbf{29.71}  \\
& GCP \cite{wu2022vivid} & 29.86 & 0.91 & 85.09 & 432.31 & 48.73 & 33.65 \\
& VCGAN \cite{Zhao_2023} & 30.2 & \textbf{0.96} & 72.17 & 2146.79  & 50.72 & 31.01 \\
& 
LatentColorization w/o TC & 30.35 & 0.92  & 18.41   & 490.89 & \textbf{44.53} &  33.84             \\ 
& LatentColorization & \textbf{30.71} & 0.93  & \textbf{17.01}   & \textbf{375.34} & 45.79 & 34.61              \\ \hline   
Sherlock Holmes Movies & DeOldify \cite{antic2019deoldify}            & -  &  -  & -  & - & \textbf{42.07} & 41.15  \\
& DeepRemaster \cite{IizukaSIGGRAPHASIA2019}        & -  & -   & -  & -  & 62.36 & 42.98 \\
& ColTran \cite{https://doi.org/10.48550/arxiv.2102.04432}            & -  & -    & - & - & 47.15 & \textbf{37.52}  \\
& GCP \cite{wu2022vivid} & - & - & - & - & 49.87 & 41.95 \\	
& VCGAN \cite{Zhao_2023} & - & - & - & - & 49.84 & 39.86 \\
& Human colorized    & -      & -      & -      & - & 48.43 & 39.78    \\
& 
LatentColorization w/o TC & - & -  & -   & - & 47.13 & 38.49        \\ 
& LatentColorization & - &  - &  -  & - & 46.24 & 41.11              \\ \hline  
Overall & DeOldify \cite{antic2019deoldify}            & 29.19  & 0.86   & 40.47  & 520.85 & 45.22 & 35.14  \\
& DeepRemaster \cite{IizukaSIGGRAPHASIA2019}        & 28.90  & 0.86   & 70.79 & 1155.24 & 55.60 & 40.13 \\
& ColTran \cite{https://doi.org/10.48550/arxiv.2102.04432}            & 29.02  & 0.83    & 64.73 & 1171.63 & 49.36 & \textbf{33.93}  \\
& GCP \cite{wu2022vivid} & 29.80 & 0.83 & 97.42 & 993.92 & 49.01 & 36.45 \\
& VCGAN \cite{Zhao_2023} & 29.03 & 0.9 & 69.98 & 1549.04 & 48.27 &  36.01 \\
& Human colorized    & -      & -      & -      & - & 48.43 & 39.78    \\
& 
LatentColorization w/o TC & 29.99 & 0.91  & 19.67   & 420.62 & 46.02 & 35.35              \\ 
& LatentColorization & \textbf{30.80} & \textbf{0.92}  & \textbf{19.64}   & \textbf{308.64} & \textbf{44.50} & 36.80              \\ \hline  
\end{tabular}
\caption{The quantitative comparisons provide a detailed evaluation of different colorization methods across various datasets. These methods include DeOldify, DeepRemaster, ColTran, GCP, VCGAN, Human Colorized, LatentColorization without Temporal Consistency and LatentColorization. The evaluation criteria encompass several metrics, including PSNR, SSIM, FID, FVD, NIQE, and BRISQUE. By comparing these metrics on individual datasets and a combined dataset (consisting of GRID, Lombard Grid, and Sherlock Holmes Movies), the study aims to assess and compare the performance of these colorization methods. This information allows for an evaluation of how LatentColorization compares to other state-of-the-art methods in various scenarios.}
\label{tab:Quantitative-Comparison}
\end{table*}

\subsection{Qualitative Analysis}

The qualitative results in Figure~\ref{fig:Comparisons} visually compare the colorization performance of different methods, including DeOldify \cite{antic2019deoldify}, ColTran \cite{https://doi.org/10.48550/arxiv.2102.04432}, DeepRemaster \cite{IizukaSIGGRAPHASIA2019}, GCP \cite{wu2022vivid}, VCGAN \cite{Zhao_2023}, LatentColorization without temporal consistency enabled, LatentColorization, and the ground truth. These comparisons are based on image sequences from the GRID \cite{cooke_martin_2006_3625687} and Lombard Grid \cite{lombardGrid} datasets. Additional qualitative results can also be seen in our appendices. This visual assessment allows for a direct comparison of how well LatentColorization performs in relation to other state-of-the-art methods. Based on the qualitative analysis of the results in Figure~\ref{fig:Comparisons}, the following conclusions can be drawn:

DeOldify \cite{antic2019deoldify} produces consistent colorizations, but they tend to appear dull and have a halo effect around the subject. ColTran generates colorful images, but it suffers from inconsistencies throughout the sequence. DeepRemaster \cite{IizukaSIGGRAPHASIA2019} provides produces dull, conservative colorizations. GCP \cite{wu2022vivid} produces colorful, consistent colorizations, but they are not faithful to the ground truth. VCGAN \cite{Zhao_2023} seems to mostly apply a blueish filter to the frames. LatentColorization w/o TC produces colorization similar to the ground truth. It is difficult to visually distinguish between LatentColorization
w/o TC, LatentColorization and the ground truth itself.
LatentColorization impressively colorizes the sequence, maintaining faithfulness to the original, vibrancy in color, and overall consistency. Overall, LatentColorization appears to outperform the other methods in terms of fidelity to the original, colorfulness, and consistency.

\subsection{Quantitative Analysis}

Quantitative evaluation is an essential aspect of assessing the quality and performance of colorization methods. It helps provide an objective measure of how well these methods perform. By evaluating colorizations both frame by frame and as a video sequence, you can gain insights into the strengths and weaknesses of each approach and determine how well they maintain consistency and quality throughout the sequence. This quantitative assessment complements the qualitative analysis and provides a more comprehensive understanding of the colorization results.

Table~\ref{tab:Quantitative-Comparison} provides a quantitative evaluation of the colorization methods, considering various image metrics. It is a useful way to compare the performance of DeOldify \cite{antic2019deoldify}, DeepRemaster \cite{IizukaSIGGRAPHASIA2019}, ColTran \cite{https://doi.org/10.48550/arxiv.2102.04432}, GCP \cite{wu2022vivid}, VCGAN \cite{Zhao_2023} , LatentColorization without temporal consistency mechanism, LatentColorization, and human colorization. By assessing metrics such as PSNR, SSIM, FID, FVD, NIQE, and BRISQUE, you can analyze the quality, similarity, and naturalness of the colorized images. This comparison enables a more data-driven and objective assessment of how well each method performs.

The results presented in Table~\ref{tab:Quantitative-Comparison} indicate that LatentColorization performs well across all of the referenced and non-referenced metrics, surpassing the state-of-the-art DeOldify \cite{antic2019deoldify} by an average of \~=18\% in terms of FVD. This performance showcases the effectiveness of LatentColorization in achieving high-quality and consistent video colorization results.

Comparing LatentColorization against human-level colorization is an important evaluation. Using non-reference image quality assessment metrics like NIQE and BRISQUE to assess the relative performance when no ground truth is available is a valuable approach. These metrics provide insights into how closely the colorization generated by LatentColorization aligns with human-expert colorization in terms of image quality.

The results in Table~\ref{tab:Quantitative-Comparison} show that LatentColorization outperforms human colorization according to NIQE and BRISQUE, which indicates that the colorizations produced by LatentColorization are of high quality when assessed using these non-reference metrics. 

The other methods also perform well on BRISQUE and NIQE scores relative to the Human Colorized version of the video. Colorization is a subjective matter, and therefore, these metrics must be paired with a user survey to evaluate the systems' performances.

\subsection{Survey}

A survey was conducted to get a more subjective view of the performance of LatentColorization. This study aimed to evaluate the difference in performance between our proposed approach, LatentColorization, and its closest competitor in our experiments, DeOldify \cite{antic2019deoldify}. Thirty-two participants were shown three sets of three videos and were asked a question on each set. Each dataset had an associated video set. The survey questions can be seen in our appendices.

For the Grid \cite{cooke_martin_2006_3625687} dataset, the participants were shown three versions of the same video taken from the dataset side-by-side. One video version had been colorized by LatentColorization, the other by DeOldify \cite{antic2019deoldify}, and the third was the ground truth. The ground truth video was labelled as such, whereas the LatentColorization and DeOldify \cite{antic2019deoldify} versions of the video were anonymous. To distinguish the LatentColorization version of the video from the DeOldify version \cite{antic2019deoldify} they were labelled with 1 and 2. After the participants had watched the videos, they were asked which video they thought was closer to the ground truth. The purpose of this question (question 1) was to differentiate in a head-to-head competition in which the colorization system was able to produce outputs which were similar to the ground truth colors of the video.

For the Lombard Grid \cite{lombardGrid} dataset, the participants were shown three versions of an example video taken from the dataset shown side-by-side. Again, one version was colorized by LatentCololorization, the other by DeOldfiy \cite{antic2019deoldify}, and the third was the ground truth. In contrast to the previous question, the ground truth video was anonymous this time, and the three videos were titled 1,2 and 3. After the participants watched the video, they were asked to rank the three videos in terms of which one looked the most realistic. Therefore, this question (question 2) acted as a visual turning test where humans were tested to see if they could tell the difference between a colorization and a ground truth video. The idea behind this is that the better the performance of the colorization system, the more difficult it should be to distinguish between the colorization system and the ground truth.

For the Sherlock Holmes dataset, the participants were shown three versions of an example video from the dataset side-by-side. One version had been colorized by LatentColorization, the other by DeOldify \cite{antic2019deoldify}, and the third was the human-colorized version. This time, the human-colorized version of the video was labelled, and the LatentColorization and DeOldify \cite{antic2019deoldify} versions were left anonymous. After the participants had watched the clips, they were asked which of the automatically colorized versions of the clip was closer to the human-colorized version. The purpose of this question (question 3) then was to determine the relative performance of LatentColorization, DeOldify \cite{antic2019deoldify} with respect to human expert colorizations. 

\begin{figure}[h]
  \centering
  \includegraphics[width=\linewidth]{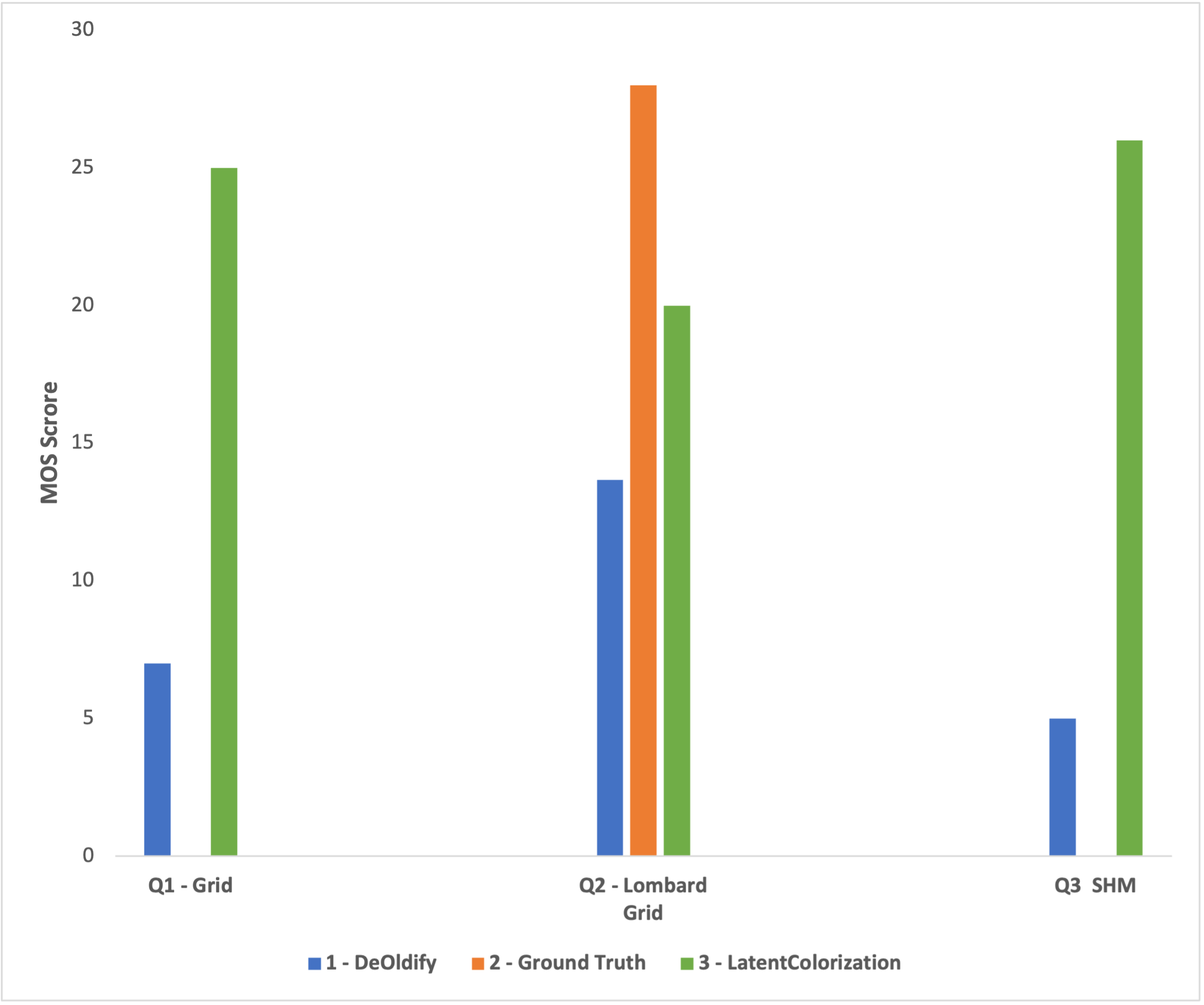}
  \caption{The graph of the results of the survey. Each group represents a particular question. The X-axis represents the Mean Opinion Score (MOS) for each question's methods. The Y axis indicates the relevant question. The color-coded bars represent each of the methods.}
  \label{fig:Survey_Results}
\end{figure}

We then collated the survey results and analysed them. The results can be seen visually in fig.~\ref{fig:Survey_Results}. The X-axis represents the Mean Opinion Score (MOS) for each question's methods. The Y axis indicates the relevant question. The color-coded bars represent each of the methods. The mean opinion score was calculated for each method for each question. For Question 1 and Question 2, the mean opinion score is simply the tally of each of the votes as it compares two methods. For Question 3, the mean opinion score is the sum of the ratings for each method divided by the number of methods.

Interpreting the graph, we can see that overall LatentColorization was preferred to DeOldify \cite{antic2019deoldify}. For question 1, DeOldify \cite{antic2019deoldify} received seven votes, and LatentColorization received 25 votes, indicating a preference for LatentColorization on this question. For question 2, the ground truth received the highest MOS score of 28.00, followed by LatentColorization at 20.00 and DeOldiy \cite{antic2019deoldify} at 13.67. Summarising this result, the ground truth was preferred most of the time, followed by LatentColorization and finally DeOldify \cite{antic2019deoldify}. For question 3, LatentColorization was chosen 26 times out of 31, indicating a strong preference for LatentColorization.

\subsection{Ablation Study}
An ablation study was undertaken to evaluate the impact of the temporal consistency mechanism on the LatentColorization system. The results for both LatentColorization and LatentColorization without temporal consistency mechanism are recorded in Table~\ref{tab:Quantitative-Comparison}. LatentColorization refers to the version of LatentColorization with the temporal consistency mechanism enabled, and LatentColorization w/o TC refers to the version of LatentColorization where the temporal consistency metric has been disabled. The results of LatentColorization and LatentColorization without temporal consistency appear similar. The main difference is that the FVD values for LatentColorization are roughly 10\% lower than LatentColorization without temporal consistency's FVD values. As a result of this observation, it can be deduced that the temporal consistency mechanism is indeed improving the video quality of the output and, therefore, ensuring temporal consistency.

\subsection{Failure Cases}

\begin{figure}[h]
  \centering
  \includegraphics[width=\linewidth]{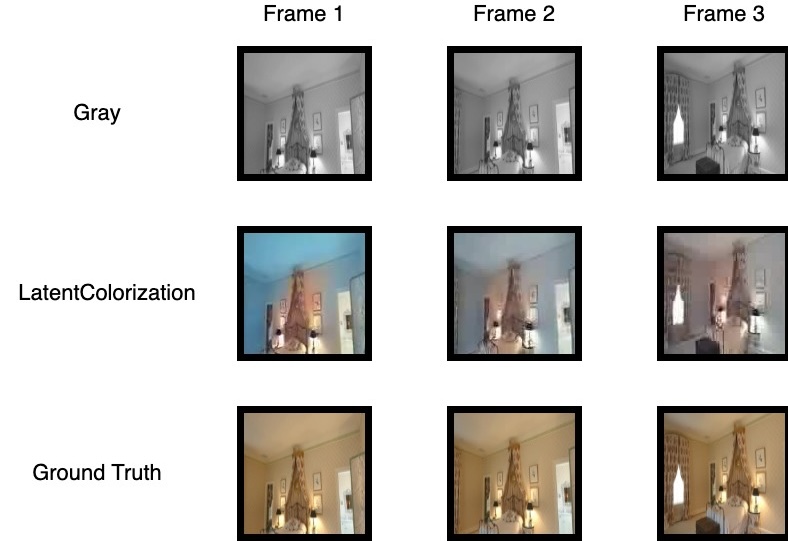}
  \caption{The comparison of three frames from the system taken from out-of-distribution data. The top row is the black-and-white version of the video, the middle frame is the output of LatentColorization, and the bottom row is the ground truth. It can be seen that LatentColorization has failed to colorize faithfully to the ground truth.}
  \label{fig:Failure_Cases}
\end{figure}

There were also instances where the system failed to colorize faithfully to the ground truth. This particularly occurred for out-of-distribution data where the videos were from a different domain than speaker videos \ref{fig:Failure_Cases}. LatentColorization fails to apply realistic colors to the bedroom scene. It initially manages to separate the walls from the bed as it colorizes the walls blue and the bed orange, see Frame 1. As time passes, see Frame 2 and Frame 3; LatentColorization tends towards a dull grey color. This indicates that LatentColorization is sensitive to the domain that the video is from, and when it does not recognize the contents of a video, it resorts to drab, dull colors.

\begin{figure*}[h]
  \centering
  \includegraphics[width=0.7\linewidth]{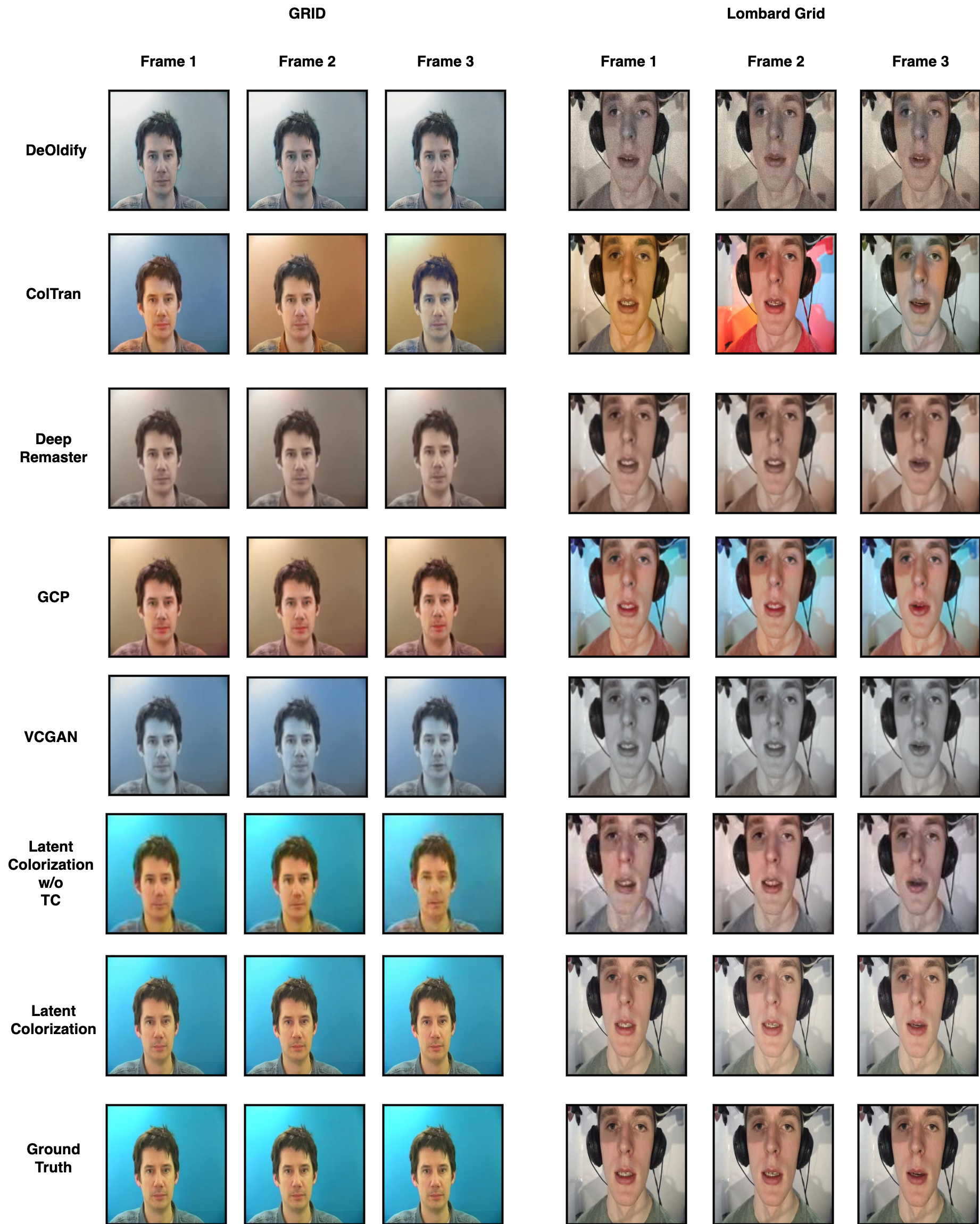}
  \caption{The qualitative comparison of colorization results from various systems, including DeOldify \cite{antic2019deoldify}, ColTran \cite{https://doi.org/10.48550/arxiv.2102.04432}, DeepRemaster \cite{IizukaSIGGRAPHASIA2019}, GCP \cite{wu2022vivid}, VCGAN \cite{Zhao_2023}, LatentColorization without the temporal consistency mechanism enabled (LatentColorization w/o TC), LatentColorization and the ground truth, for both the GRID \cite{cooke_martin_2006_3625687} dataset (left) and the Lombard Grid \cite{lombardGrid} dataset (right) is shown. In the GRID \cite{cooke_martin_2006_3625687} dataset, DeOldify's \cite{antic2019deoldify} colorization, depicted in the first row, exhibits desaturated colors and a halo effect around the subject. ColTran \cite{https://doi.org/10.48550/arxiv.2102.04432}, in the second row, produces more colorful results but lacks consistency throughout the sequence. DeepRemaster \cite{IizukaSIGGRAPHASIA2019} produces dull, conservative colorizations. GCP \cite{wu2022vivid} produces colorful, consistent colorizations, but they are not faithful to the ground truth. VCGAN \cite{Zhao_2023} produces drab, monotone colorizations.  LatentColorization w/o TC produces colorization similar to the ground truth. It is difficult to visually distinguish between  LatentColorization w/o TC, LatentColorization and the ground truth itself. The ground truth, represents the original color frames. Similar observations can be made for the Lombard Grid \cite{lombardGrid} dataset. These visual comparisons demonstrate that LatentColorization consistently delivers colorization results that closely match the original colors, making it a promising technique for automatic video colorization tasks.}
  \label{fig:Comparisons}
\end{figure*}

\section{Discussion}
\label{sec:discussion}

In this section, we discuss our model's results compared to other approaches from the field and an overview highlighting the main limitations associated with our approach. 

\subsection{Model Comparisons}

The comparison between LatentColorization and non-autoregressive models like ColTran \cite{https://doi.org/10.48550/arxiv.2102.04432} provides insights into the importance of the autoregressive nature of the system in the context of video colorization. Fig~\ref{fig:Comparisons} demonstrates the difference in consistency between the two approaches. The frames colorized by LatentColorization appear more consistent throughout the video sequence, while those generated by ColTran \cite{https://doi.org/10.48550/arxiv.2102.04432} exhibit more variation. This suggests that the autoregressive nature of LatentColorization, where each frame is conditioned on the previous ones, plays a role in maintaining temporal consistency and ensuring that the colorization is coherent across the entire video. In contrast, non-autoregressive approaches like ColTran \cite{https://doi.org/10.48550/arxiv.2102.04432} may struggle to achieve the same level of consistency in colorized sequences.

The qualitative assessment of the colorizations in Fig~\ref{fig:Comparisons} highlights the differences in colorfulness among LatentColorization, and DeOldify \cite{antic2019deoldify}. LatentColorization produces colorful results. In contrast, DeOldify \cite{antic2019deoldify} appears grey, suggesting that it may suffer from a lack of color diversity. This observation is consistent with the idea that GANs, which DeOldify \cite{antic2019deoldify} is based on, can be susceptible to mode collapse, where they produce limited and less diverse color variations. This observation also correlated with the survey results where LatentColorization was preferred to DeOldify \cite{antic2019deoldify} 80\% of the time.

\textbf{DeepRemaster \cite{IizukaSIGGRAPHASIA2019} Vs LatentColorization}: DeepRemaster \cite{IizukaSIGGRAPHASIA2019} has struggled with the colorization of this material and has resorted to very bland, dull colors, unlike LatentColorization. 

\textbf{GCP \cite{wu2022vivid} Vs LatentColorization}: it can be seen that LatentColorization is closer to the ground truth than GCP \cite{wu2022vivid}. GCP has produced colorful output, but it is different in color from the ground truth. It has not succumbed to the mode collapse of its GAN-based architecture, especially on the Lombard Grid \cite{lombardGrid} dataset. This could potentially be a result of its retrieval mechanism.

\textbf{VCGAN \cite{Zhao_2023} Vs LatentColorization}: it can be seen that LatentColorization is closer to the ground truth than VCGAN \cite{Zhao_2023}. VCGAN has produced a blue filter type effect on the frames.

\textbf{LatentColorization Vs LatentColorization without temporal consistency}: has been investigated in the ablation study. Essentially, it is difficult to visually differentiate between the two, and the main difference can be seen quantitatively in their relative FVD scores.

The quantitative evaluation, as shown in Table~\ref{tab:Quantitative-Comparison}, indicates that LatentColorization achieved scores on the NIQE and BRISQUE metrics that are close to human-level colorization. In summary, these results suggest that LatentColorization, in this experiment, is comparable to human-level colorization in terms of the assessed quality metrics. This highlights the effectiveness of the LatentColorization method in generating high-quality colorized videos. This evaluation also correlates with our survey, where LatentColorization received a higher preference from the subjects than DeOldify \cite{antic2019deoldify}. The survey also shows a tendency of the users to prefer the ground truth videos over both LatentColorization and DeOldify \cite{antic2019deoldify}.

\subsection{Limitations}

One of the main limitations of our approach is that the datasets we use are specific to speaker videos, causing our model to perform more poorly on out of domain data. We intend to address this in our future work by training our model on a more diverse dataset capturing a wide range of scenarios. 

Our model also exhibits poor performance when it compares to inference speed. For instance, colorizing a five-second clip at fifty diffusion steps takes roughly one hundred and fifty seconds on an Nvidia 2080 with 8GB of VRAM. One of the main drawbacks to a diffusion model-based system is its inference time, as the model must sample each frame equal to the number of diffusion steps chosen. Real-time colorization is beyond the scope of this work, but generally, real-time is not a requirement in applications.

Ethical concerns which must also be considered. Two main worries associated with this type of technology are potential misuse and bias. Defining the potential misuse of colorization systems is a difficult task with various nuances. Opponents of colorization believe that it is unnecessary and defaces the original work. Proponents retort that it makes the material more approachable to wider audiences \cite{vertigo}.

In addition to the ethical considerations surrounding the potential issue of these systems, there are also concerns regarding the bias of these systems \cite{stapel2022bias}. Through experimentation it has been found that these systems can be susceptible to tending towards outputs which are similar to the data that they were trained on. As datasets can be biased, the models can also inherit this bias and therefore output inaccurate results. In colorization systems, this can present itself in such manners as incorrect skin colors or incorrect color uniforms which may give a distorted view of history. 

\section{Conclusion}
\label{sec:conclusion}

In conclusion, our work demonstrates the effectiveness of diffusion-based models, particularly the LatentColorization method, in achieving results comparable to the state of the art across multiple datasets. Notably, the system performs comparably to human-level colorization on the 'Sherlock Holmes Movie' dataset, indicating its practical significance and the potential for application-specific video colorization. The use of a latent diffusion model and the incorporation of a temporally consistent colorization approach contribute to realistic and convincing colorization results, making the process more accessible and reducing the reliance on traditional human-driven colorization methods. This research provides insights into the potential of diffusion models for video colorization and opens up opportunities for further developments in this field.

\section{Future Work}
\label{sec:future_work}

Expanding on our research, adapting the system to work with various video styles, types, and content would be a promising direction for future work, enabling a broader assessment of the approach's applicability in the context of general video colorization. These endeavors will further enhance the practicality and versatility of our research in the field of automatic video colorization.

\bibliographystyle{unsrt}
\bibliography{LatentColorization:_Latent_Diffusion_Based_Video_Colorization_V2}

\begin{IEEEbiography}[{\includegraphics[width=1in,height=1.25in,clip,keepaspectratio]{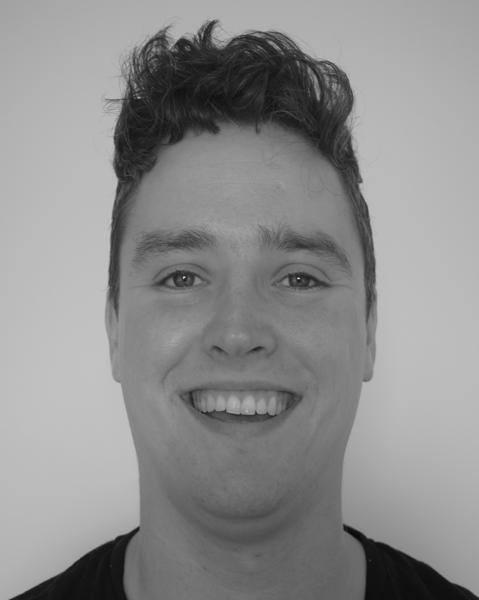}}]{Rory Ward} (Member, IEEE) received the bachelor's degree in electronic and
computer engineering from the National University of Ireland Galway, in 2021. He is pursuing his PhD at the University of Galway, where he is sponsored by the Centre for Research Training in Artificial Intelligence. His research interest includes automatic visual media colorization.
\end{IEEEbiography}

\begin{IEEEbiography}[{\includegraphics[width=1in,height=1.25in,clip,keepaspectratio]{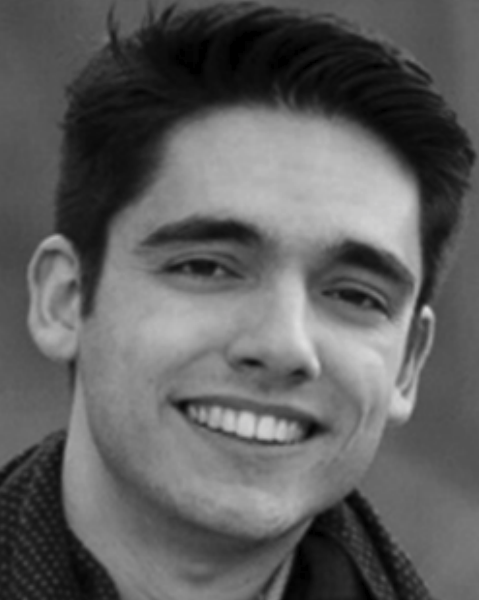}}]{DAN BIGIOI} (Graduate Student Member, IEEE)
received a bachelor's degree in electronic and
computer engineering from the National University of Ireland Galway in 2020. Upon graduating, he worked as a Research Assistant at NUIG
studying the text-to-speech and speaker recognition methods under the DAVID (Data-Center
Audio/Visual Intelligence on-Device) Project.
Currently, he is working on his PhD at NUIG,
sponsored by D-REAL and the SFI Centre for
Research Training in Digitally Enhanced Reality. His research interests
include novel deep learning-based techniques for automatic speech dubbing
and discovering new ways to process multimodal audio/visual data.
\end{IEEEbiography}

\begin{IEEEbiography}[{\includegraphics[width=1in,height=1.25in,clip,keepaspectratio]{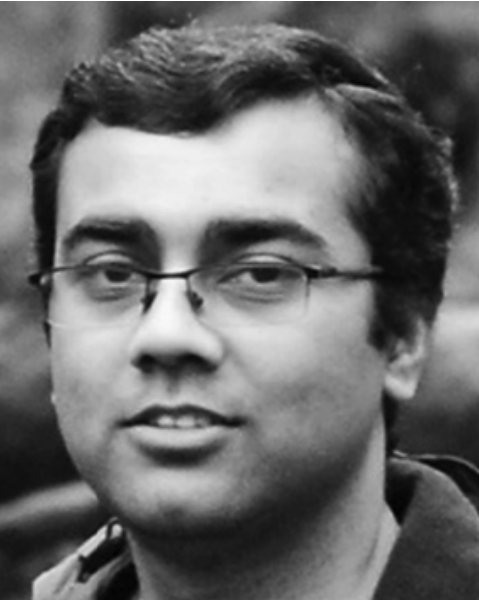}}]{SHUBHAJIT BASAK} received the B.Tech. degree
in electronics and communication engineering
from the West Bengal University of Technology,
India, in 2011, and the M.Sc. degree in computer
science from the National University of Ireland
Galway, Ireland, in 2018, where he is currently
pursuing a Ph.D. degree in computer science.
He has over six years of industrial experience
as a Software Development Professional. He is
also with FotoNation/Xperi. His research interest
includes deep learning tasks related to computer vision.
\end{IEEEbiography}

\begin{IEEEbiography}[{\includegraphics[width=1in,height=1.25in,clip,keepaspectratio]{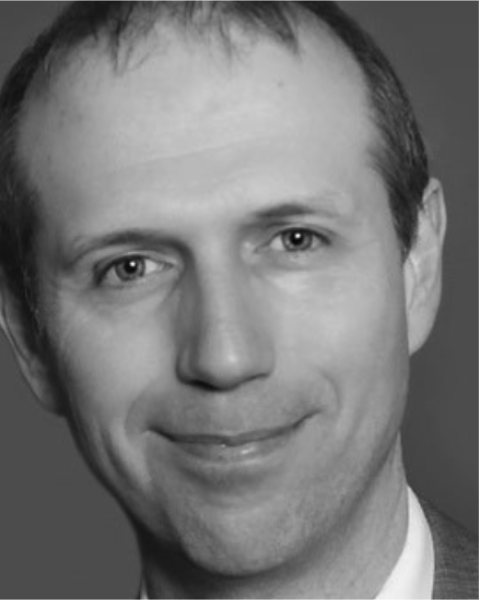}}]{JOHN G. BRESLIN}(Senior Member, IEEE) is a Personal Professor (Personal Chair) in Electronic Engineering with the College of Science and Engineering, University of Galway, Ireland, where he is the Director of the TechInnovate/AgInnovate programs. He has taught electronic engineering, computer science, innovation, and entrepreneurship topics during the past two decades. He is associated with two SFI Research Centers, where he is a Co-Principal Investigator with Insight (Data Analytics) and a Funded Investigator with VistaMilk (AgTech). He has written more than 250 peer-reviewed academic publications with a h-index of 50 and 10000 citations. He has co-authored the books Old Ireland in Colour 1, 2 and 3, The Social Semantic Web, and Social Semantic Web Mining. He has co-created the SIOC framework, which is implemented in hundreds of applications (by Yahoo, Boeing, and Vodafone) on at least 65,000 websites with 35 million data instances. He received various Best Paper Awards from conferences and journals.

\end{IEEEbiography}

\begin{IEEEbiography}[{\includegraphics[width=1in,height=1.25in,clip,keepaspectratio]{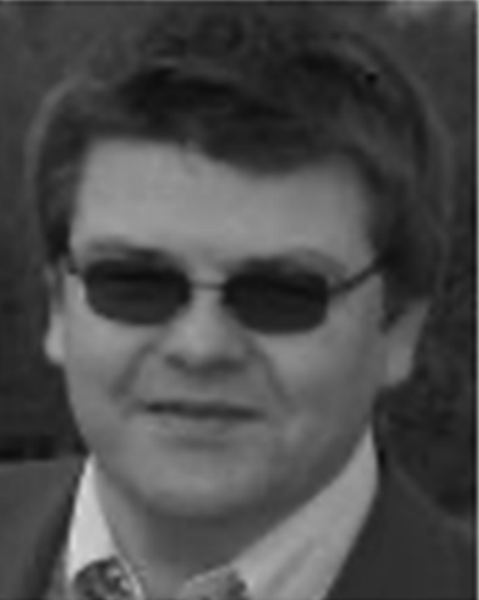}}]{PETER CORCORAN} (Fellow, IEEE) is currently holding the Personal Chair of Electronic Engineering with the College of Science and Engineering, University of Galway.
He was the Co-Founder of several start-up companies, notably FotoNation (currently the Imaging
Division, Xperi Corporation). He has more than
600 cited technical publications and patents, more
than 120 peer-reviewed journal articles, 160 international conference papers, and a co-inventor on
more than 300 granted U.S. patents. He is an IEEE Fellow recognized for
his contributions to digital camera technologies, notably in-camera red-eye
correction and facial detection. He is also a member of the IEEE Consumer
Technology Society for more than 25 years and the Founding Editor of IEEE
Consumer Electronics Magazine.
\end{IEEEbiography}
\EOD

\appendix\section{Additional Experiments}\label{appendix:additional_experiments}

\begin{figure*}[h]
  \centering
  \includegraphics[width=0.8\linewidth]{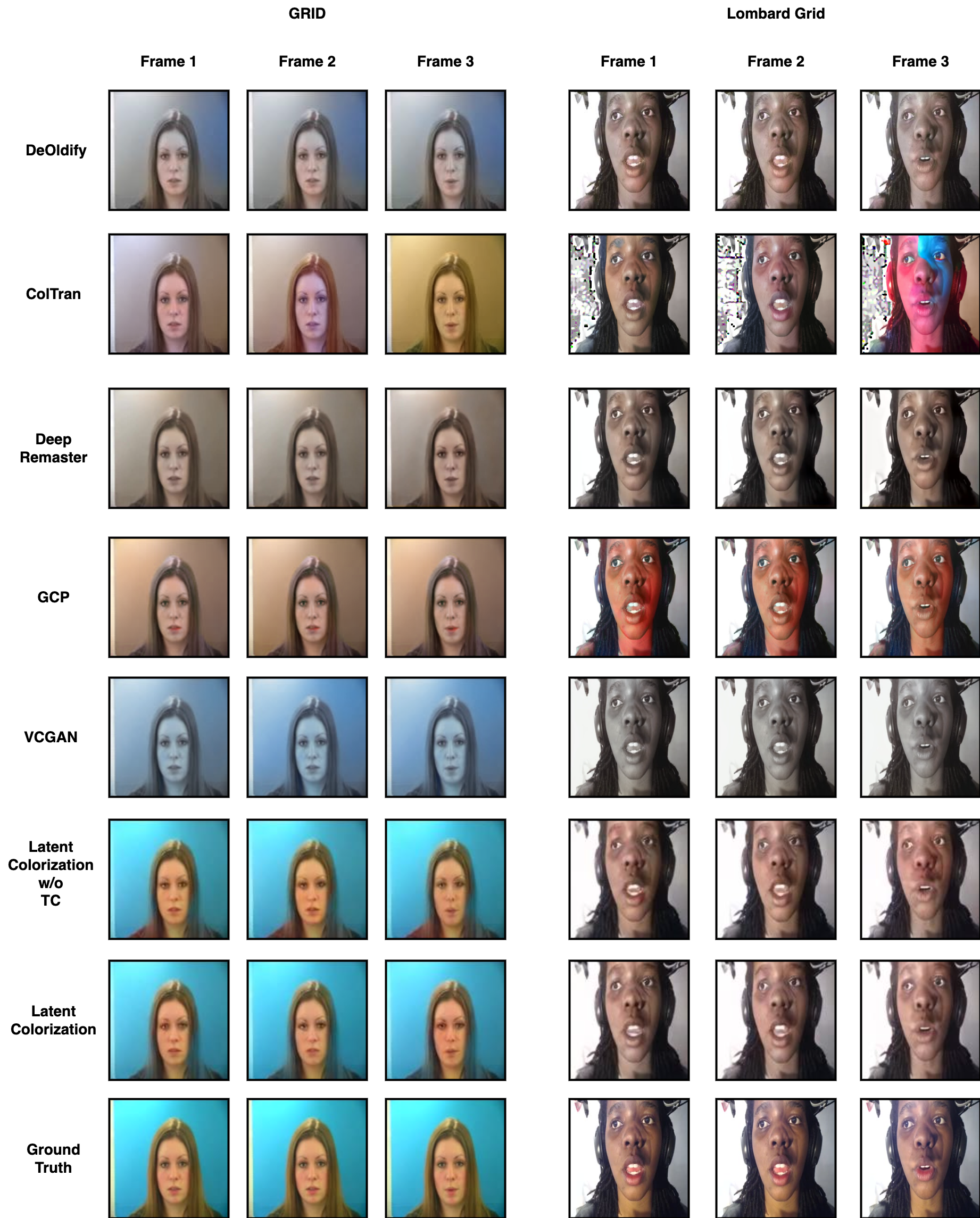}
  \caption{The qualitative comparison of colorization results from various systems, including DeOldify \cite{antic2019deoldify}, ColTran \cite{https://doi.org/10.48550/arxiv.2102.04432}, DeepRemaster \cite{IizukaSIGGRAPHASIA2019}, GCP \cite{wu2022vivid}, VCGAN \cite{Zhao_2023}, LatentColorization without the temporal consistency mechanism enabled (LatentColorization w/o TC), LatentColorization and the ground truth, for both the GRID \cite{cooke_martin_2006_3625687} dataset (left) and the Lombard Grid \cite{lombardGrid} dataset (right) reveals differences in their performance.}
  \label{fig:Comparisons_2}
\end{figure*}

\begin{figure*}[h]
  \centering
  \includegraphics[width=0.8\linewidth]{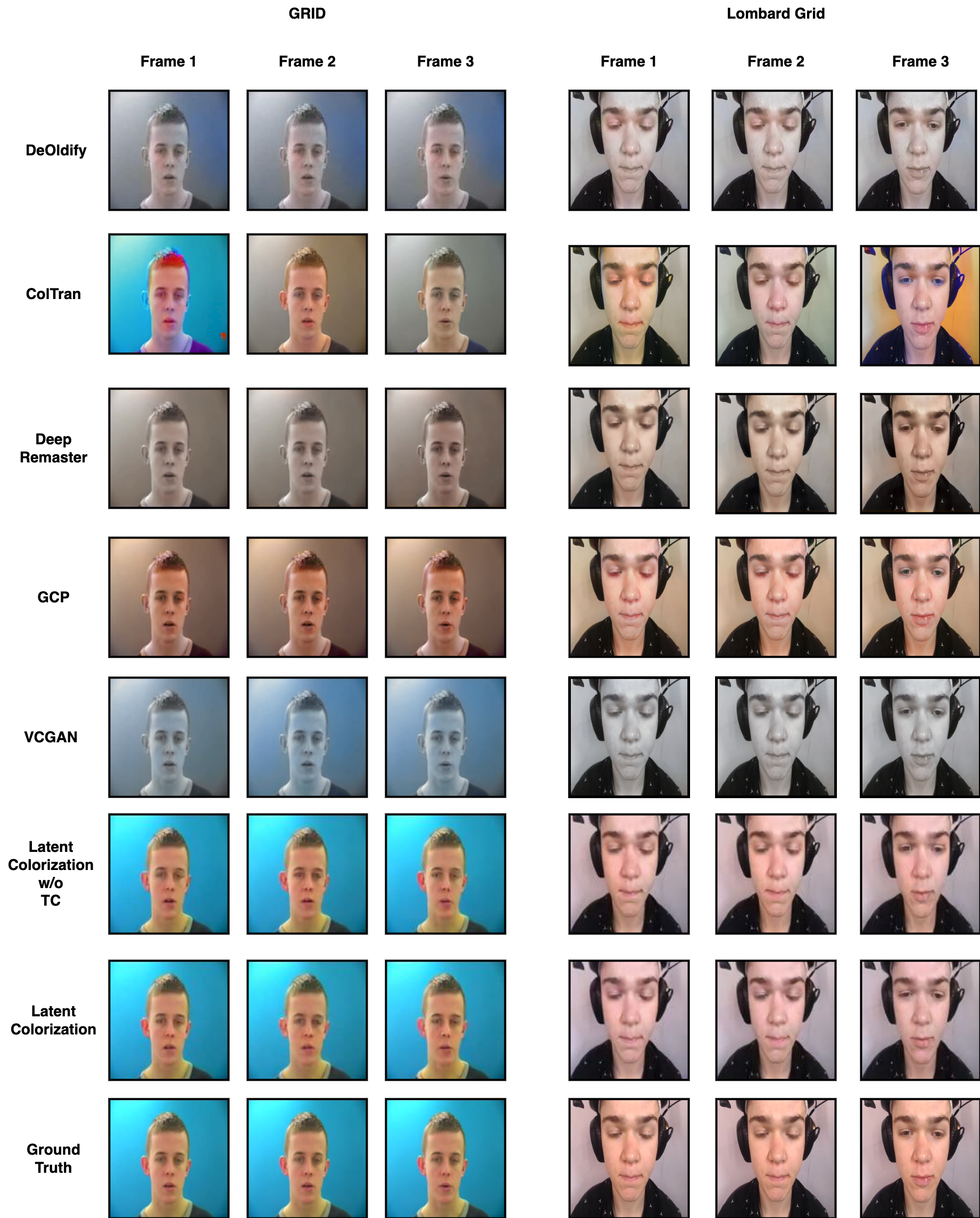}
  \caption{The qualitative comparison of colorization results from various systems, including DeOldify \cite{antic2019deoldify}, ColTran \cite{https://doi.org/10.48550/arxiv.2102.04432}, DeepRemaster \cite{IizukaSIGGRAPHASIA2019}, GCP \cite{wu2022vivid}, VCGAN \cite{Zhao_2023}, LatentColorization without the temporal consistency mechanism enabled (LatentColorization w/o TC), LatentColorization and the ground truth, for both the GRID \cite{cooke_martin_2006_3625687} dataset (left) and the Lombard Grid \cite{lombardGrid} dataset (right) reveals differences in their performance.}
  \label{fig:Comparisons_3}
\end{figure*}

\appendix\section{Survey}\label{appendix:survey}

\begin{figure*}[h]
  \centering
  \includegraphics[width=1\linewidth]{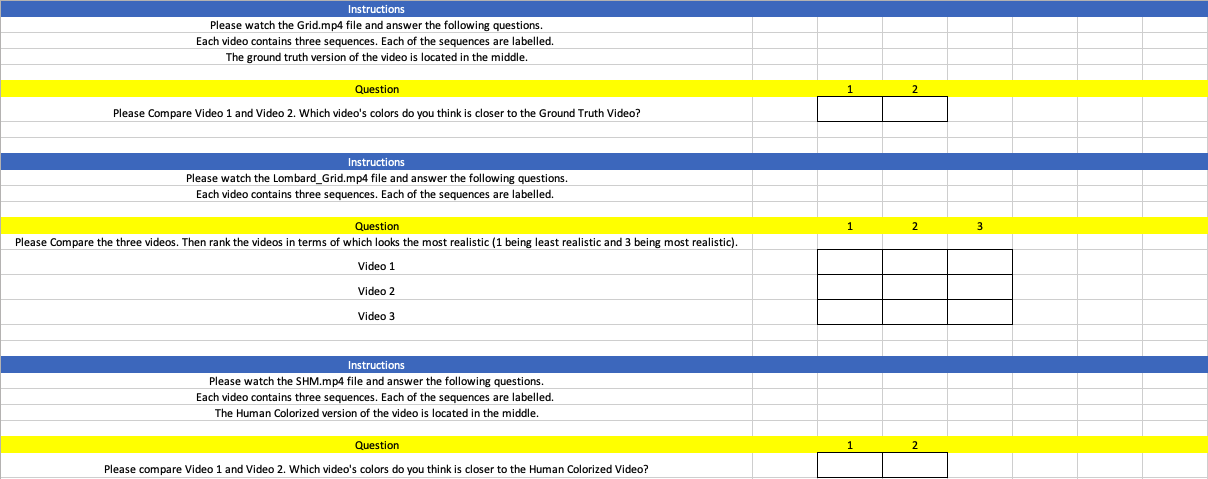}
  \caption{The survey questions asked. Question 1 compares LatentColorization to DeOldify \cite{antic2019deoldify} on the Grid \cite{cooke_martin_2006_3625687} dataset. Question 2 compares LatentColorization, DeOldify \cite{antic2019deoldify} and the ground truth on the Lombard Grid \cite{lombardGrid} dataset. Question 3 compares LatentColorization, and DeOldify \cite{antic2019deoldify} on the Sherlock Holmes Movie dataset.}
  \label{fig:Survey}
\end{figure*}

\end{document}